\def\year{2018}\relax
\documentclass[letterpaper]{article} 
\usepackage{aaai18}  
\usepackage{times}  

    \usepackage{ulem}  

\usepackage{helvet}  
\usepackage{courier}  
\usepackage{url}  
\usepackage{graphicx}  
\begin{document}
%
\title{Using experimental game theory to transit human values to ethical AI }
\author{Yijia Wang$^1$, Yan Wan$^2$ and Zhijian Wang$^3$\\
$^1$ School of Mathematical Sciences, University of Electronic Science and Technology of China, Chengdu 611731, China\\
$^2$ School of Economics \& Management, Beijing University of Posts and telecommunications, Beijing, China\\
$^3$ Experimental Social Science Laboratoy, Zhejiang University, Hangzhou 310058, China\\
}
\maketitle
\begin{abstract}
Knowing the reflection of game theory and ethics, we develop a mathematical representation to bridge the gap between the concepts in moral philosophy (e.g., Kantian and Utilitarian) and AI ethics industry technology standard (e.g., IEEE P7000 standard series for Ethical AI). As an application, we demonstrate how human value can be obtained from the experimental game theory (e.g., trust game experiment) so as to build an ethical AI. Moreover, an approach to test the ethics (rightness or wrongness) of a given AI algorithm by using an iterated Prisoner's Dilemma Game experiment is discussed as an example. Compared with existing mathematical frameworks and testing method on AI ethics technology, the advantages of the proposed approach are analyzed.
\end{abstract}

\section{Introduction}
Pace of AI development brings up the general worriment
on human life (e.g., physical safety, mental happiness)
and social impacts (e.g., workforce displacement, economics inequality, etc) \cite{Allen2006Why}.
Such worriment is not unreasonable.
Because the outcome of AI technology appears unpredictable,
which significantly differs from our daily productions
whose outcome is deductive, predictable and controllable based on existed physics, chemistry, biology, mathematics science and engineering.

\begin{figure}[tbh]
  \centering
      \includegraphics[scale = 0.30]{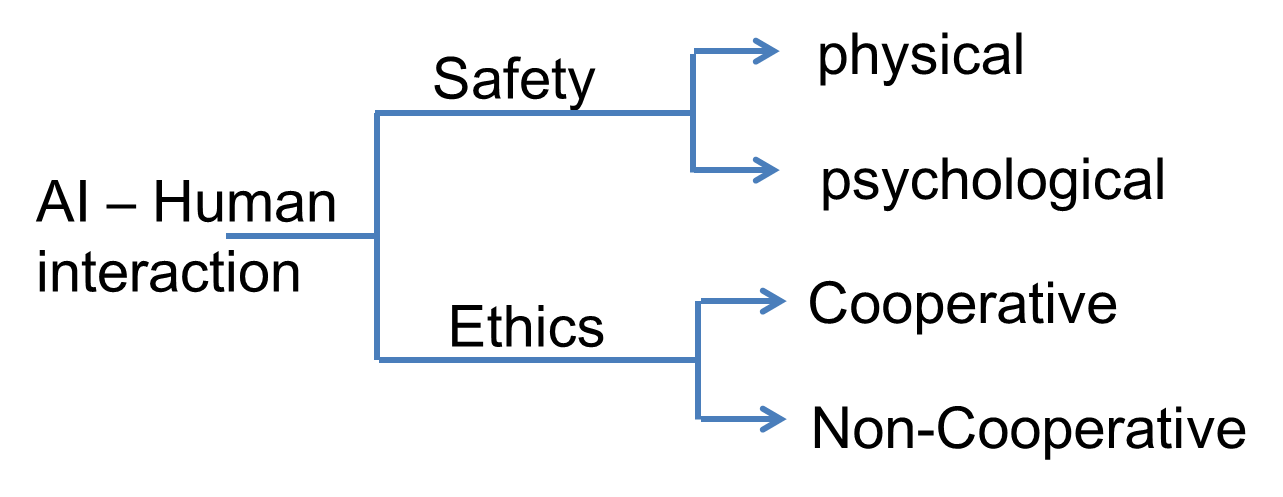}  \\
  \caption{AI ethics issues in AI-Human interaction.}\label{fig:TreeChart1}
\end{figure}

On AI and human interactions, issues can be presented in Fig. \ref{fig:TreeChart1}. AI safety, which can be categrised as physical safety and psychological safety, is well understood \cite{MITsafety2017} and related industry technology standards have also been exemplified  established (e.g., ISO 10218-1:2011, ISO 15066). However, the ethical issues on AI decision making are remained. The questions can be expressed as how to embed human values into
AI, or how to align AI behaviors with human value \cite{Raja2017The}. More strictly speaking, methodologies to guide AI ethics research and design are needed.

Methodology for AI ethics research and design is not blank.
Ongoing AI ethic industry technology standards IEEE P7000 is based on Value Sensitive Design method \cite{Friedman2013Value}, and IEEE P7010 is based on happiness index methods. We notice that, all the methods applied on AI ethics are inductive. These methods appear not structural and lacking of convincing, which could lead to conflict of the rules, overload or blind spot, referring to \cite{sowa1999knowledge}. Logically, methods can be either inductive or deductive, but till now, there has been little research going on deductive method for AI ethics. So, for AI to be aligned with human value, a rigorous method,which has a inclusive and clear mathematical paradigm, and is theoretical computable, experimentally testable and industry technology conductible, is desired.

In this paper, we introduce a mathematical presentation for AI-human interaction, which can tolerate the conflict of Kantian and Utilitarian moral philosophy, and turn the interactions  computable when human values are included. More importantly, we demonstrate that the human value in non-cooperate game condition can be taken from laboratory for AI ethics behavior design. We show a technology to
test the ethics (rightness or wrongness) of a given AI algorithm
quantitatively. Comparison with related literatures
about the science and technology on AI ethics issues,
as well as the further researches will be discussed.

\section{Technology Note}

 Aiming at developing practical approach to
 transit human value into AI industry production,
 a brief glossary of terminology is summarized as follows, instead of describing the concepts in moral philosophy, game theory or experimental behavior science in a very detailed way.

\subsubsection{AI Ethics}
AI Ethics, in this paper, is about the 'rightness or wrongness' of an AI agent behaviors when interacting with human being (in AI-human interaction).
Here, the AI agent is an individual
 (e.g., autonomous cars, robotics, drones, financial agents, an so on),
 and its behaviors are based on its solo decisions.
The 'rightness or wrongness', as a concept in moral philosophy, is defined by Kantian and Utilitarian simultaneously.

\subsubsection{Kantian and Utilitarian} focus on the input and output of an AI-human interaction, respectively. Mathematically \cite{Osborne1994A}, the input can be presented in \emph{strategy space} as the vector of an action, and the outcome can be presented in outcome space as the vector of the results of the interaction.
Strategy space
is the space of all possible actions (behaviors) which an AI agent or a human being
can apply in the interaction.
The \emph{outcome space} presents the payoff (physical and mental reward and cost) of each subject involved in the strategy interaction. The structure of the outcome space can be multi-dimension and various presentation (total social outcome, fairness, etc).

 \subsubsection{Experimental game theory} is an inter-disciplinary area of game theory and human behavior experiments, studying human decision-making. In non-cooperation game, the force (incentive, due to Kantian, e.g., fairness behavior in dictator game, altruism and punishment in public good games) drives the behaviors deviated from Utilitarian (rational choice) has been extensively studied.

\subsubsection{Human value alignment}
 in this report, we follow IEEE global initiative ethical design.
Quantitatively, the human value is specified as the Kantian value, which is beyond the rational Utilitarian value in this paper, and is expected to be aligned by AI in industry technology standard.
The main point of this paper is to accommodate Utilitarian and Kantian with a computational mathematical paradigm, with which we can use human behavior data to transit the human value to ethical AI.

\section{Mathematical representation}

AI Ethics, as well as moral philosophy or common goodness willing, needs a mathematical representation before it could be conducted practically in the AI production life-cycle. Or, to a large extent, the investigation or discussion is not a technical question anyways and let us fall into chaos among game theory,
morality, psychology, personal dogmatism, etc.
Our developing the mathematical presentation for AI ethics is not along, and 
comparison between ours and the previous \cite{Conitzer2017Moral,Letchford2008An} will be introduced later.

According to the AI Ethics definition, we can formulate the behavior interactions and the outcome as
\begin{equation}\label{eq1}
   \textbf{S}^a_i \otimes \textbf{S}^b_j \rightarrow \textbf{O},
\end{equation}
in which, $\textbf{S}^a_i$ indicates an AI \textbf{a}gent applies strategy $i$ in its strategy space $\textbf{S}^a$,  $\textbf{S}^b_i$ indicates a human \textbf{b}eing applies  strategy $j$ in its strategy space $\textbf{S}^b$, and  \textbf{O} indicates the outcome space. This kind of representation has been used to investigate the reflection of game theory and ethics, see reference \cite{Kuhn2004Reflections} and \cite{Conitzer2017Moral}. In such a representation, if one side's strategy is fixed (supposing strategy $j$ in $\textbf{S}^b$ is given and known by AI agent), then the decision making turns to optimization question, and $\textbf{S}^a_i$ determines the outcome,  which is the basis of evaluation of the ethics of AI.

\begin{figure}[tbh]
  \centering
      \includegraphics[scale = 0.24]{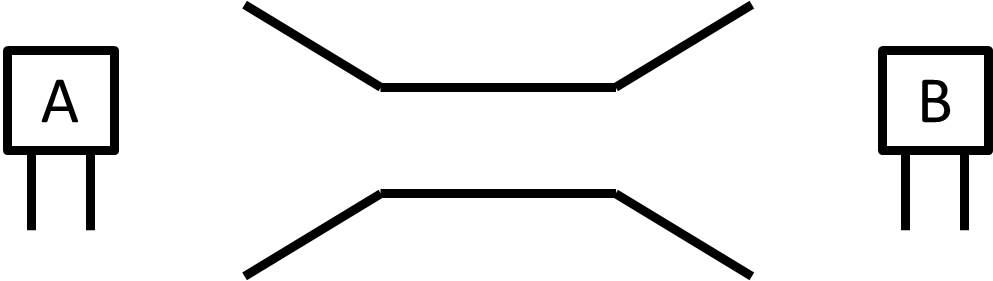}  \\
  \caption{Wait or pass game. This is a strategy interaction game, in which an AI agent (A) meets a human being (B) on a narrow bridge. }\label{fig:bridge1}
\end{figure}

\subsubsection{An example}Using a Wait or Pass dilemma game, we illustrate the meaning of the concepts in technology note and the formula above. Figure 2 demonstrates a situation, in which an AI agent meets a human being agent on a narrow bridge, but is too narrow for two agents to pass simultaneously.  At this moment, for both AI agent and human agent, the optional behaviors (or strategies) can only be wait or pass, presented as
\begin{eqnarray}\label{eq2}
  \textbf{S}^a  &=&  \{{wait, pass}\}, \\
  \textbf{S}^b   &=& \{{wait, pass}\}
\end{eqnarray}

At first stage, Utilitarian but not Kantian, we ignore the Kantian value (denoted as \textbf{K}, indicates rightness or wrongness) of an action by setting \textbf{K$_{wait}$} = \textbf{K$_{pass}$} = 0, but only take the outcome into account as a rational individual in traditional game theory.
If both use 'wait' strategy, both agent lost 1 (denoted as gain -1) unit of utility,
if both use 'pass' strategy, both agent lost 2 (denoted as gain -2) unit of utility.
And If one uses 'wait' strategy and the opponent uses 'pass' strategy, the 'wait' strategy user gains -1, and the 'pass' strategy user gains 1. So, the outcome space \textbf{O} includes 4 elements, as the result from 2 (elements belongs to $\textbf{S}^a$) times (denoted as $\otimes$) 2 (total elements belongs to $\textbf{S}^b$). More visually, the Values in the outcome space of this game is illustrated in Table \ref{Table:01}. \\

\begin{table}[thp] 
\caption{Values presented in the outcome space$^\textbf{*}$}\label{Table:01}
\centering
\begin{tabular*}{7.9cm}{@{\qquad}c@{\qquad}c@{\qquad}c@{\qquad}c@{\qquad}}
\hline
AI agent  & Human & AI agent    & Human \\
strategy       & strategy    & reward & reward \\
\hline
wait & wait & -1~+  \textbf{K$_{wait}$}  & -1 \\
  wait & pass & -1~+  \textbf{K$_{wait}$} & ~1 \\
  pass & wait & ~1~+  \textbf{K$_{pass}$} & -1 \\
  pass & pass & -2~+  \textbf{K$_{pass}$} & -2 \\
\hline
\multicolumn{4}{p{7.6cm}}{\scriptsize $^*$ unit of reward is ignored. }
\end{tabular*}
\end{table}

\subsubsection{Computable AI ethics}
In game theory view, so-called AI ethical behavior constraint is to limit the AI strategy in $\textbf{S}^a$ by evaluating the $\textbf{O}$, which appears to be a Utilitarian method and also be computable. Alternatively, in Kantian view, an AI agent should not select the ethically wrong strategies in $\textbf{S}^a$, which is a obligation. The Utilitarian and the Kantian could lead to moral dilemma. As real ethical approach, we will see that, although various environments and ethical standard can lead to different results, our approach remains computable.

\begin{enumerate}
  \item  Totally prioritizing human well-being rule can be implemented on the outcome space, by seeking the maximum human reward shown in 4th column in Table \ref{Table:01}. It is obvious that, when AI agent uses 'wait' strategy, whatever human will select, human's gain will be better. Then such an AI is considered to be ethical. This scheme, in fact, turns a non-cooperative game to a cooperative game, and makes it computable.
  \item Prioritizing social well-being (or Pareto efficient), and attaching further consideration to social responsibility, the algorithm will seek the maximum sum of AI agent and human reward shown in 3rd and 4th column in Table \ref{Table:01}. It can be seen that, in this case, there is no pure solution for AI strategy in this condition.
  \item Partly prioritizing human well-being can be implemented on the outcome space, by assigning weight of a prioritizing parameter ($\beta >$ 1) to the human reward on the fourth column in Table \ref{Table:01}. And then, AI regards its reward as the sum of the 3rd column plus $\beta$ times 4th column, consequently the solution for AI will be wait. If an industry standard for AI ethics wants to emphasize the prioritizing, the question turns to how to determine the $\beta$ value. Once $\beta$ is determined, the AI ethics is computable.
  \item Totally Kantian requires that only the action itself has its value (rightness or wrongness).
  Supposing 'to be modest' is right, we can set  \textbf{K$_{wait}$} = $\infty$ in Table \ref{Table:01}.
  Moral relativism follows Kantian requirement but set  \textbf{K$_{wait}$} to a finite value in Table \ref{Table:01}.
  Having Kantian \textbf{K} value included in the outcome space, the computing technology in game theory can be applied.
\end{enumerate}

In summary, in the first two (1, 2) conditions, the solution for AI ethics becomes the solution for optimization problem, the solution for a cooperative game. And in the last two (3,4) conditions, the solutions becomes the Nash equilibria, the solution of a non-cooperation game (e.g., an auto financial algorithm stock robot competes with its human opponent, or an gas station robot competes with its neighbor gas station in a price war). Having the solution,
we could make a quantitative
  assessment of how probable
of the action of AI agent being ethical.

As it can be seen, in a real life condition, the strategy space can be large, and the values are not clear. Here the presentation provides a framework, which makes AI ethics negotiable between its stakeholder.

\section{Take human value from trust game}

Game theory, originally, studies the fully rational (Utilitarian) behavior in agents interaction.
However in experiments, human subjects' behaviors deviate significantly from the fully rational ones, which is exactly the ethical behavior or the goodness of human being that is expected to be captured in the experiment data.
Since the experiments can provide statistical results in controlled environment, the value can be taken quantitatively.
In this section, using trust game experiment as an example, we propose an approach to take the human value --- Kantian ---from experiments, which can be transited to ethical AI in technology design.

\subsubsection{Trust game}
In standard trust games, no trust is expected by fully rational hypothesis, but in experiments trust could generates a potential gain.
The following two person game shown in Figure 3 is commonly studied by experimental economists, and summarized by \cite{Smith2014Fair} in a variety of forms.
 Person 1 chooses to either (a) end the game and each person earning with  \$10 or (b) forego his sure \$10 and turn the decision making to Person 2. If Person 1 chooses (b), then Person 2 decides between (a') the experimenter paying her \$25 and Person 1 \$15 or (b') the experimenter paying her \$40 and sending Person 1 on his way with nothing by way of the outcome from the interaction in this game. Applying the concept of subgame perfect equilibrium, in backward induce, a' and b are dominated strategies, the Nash equilibrium solution is Person 1 chooses a, and each person earning with  \$10.

\subsubsection{Experiment results ---}
In the laboratory, referring to the summary by \cite{Smith2014Fair}, the replicable facts from three different studies are that in 98 (Person 1) first movers, 52 choose (a) and 46 choose (b), and that of the 46 (Person 2) second movers who have the opportunity to make a decision, 31 (67\%) choose (a') and 15 (33\%) choose (b'). On average, a Person 1 gains about \$10 and a Person 2 gains about \$20.


\begin{figure}[tbh]
  \centering
      \includegraphics[scale = 0.21]{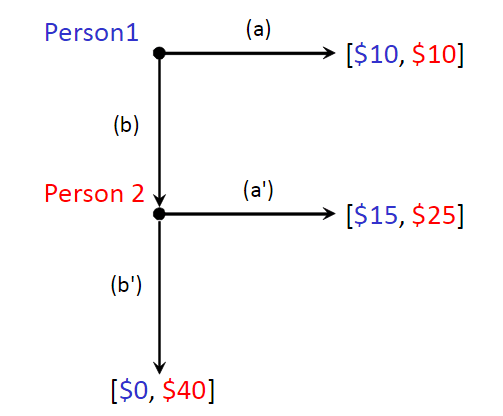}
  \caption{A trust game in extensive form }\label{fig:trustgame}
\end{figure}

\subsubsection{Taking the human value from data ---}
So-called human value, in this paper, is defined as the value that drive human decisions to deviate from Utilitarian (fully rational) behaviors in the game.
Despite various explanations towards that, here we just need to take the value out numerically for AI to learn or for specifying AI ethical behaviors.

\begin{figure}[tbh]
  \centering
      \includegraphics[scale = 0.26]{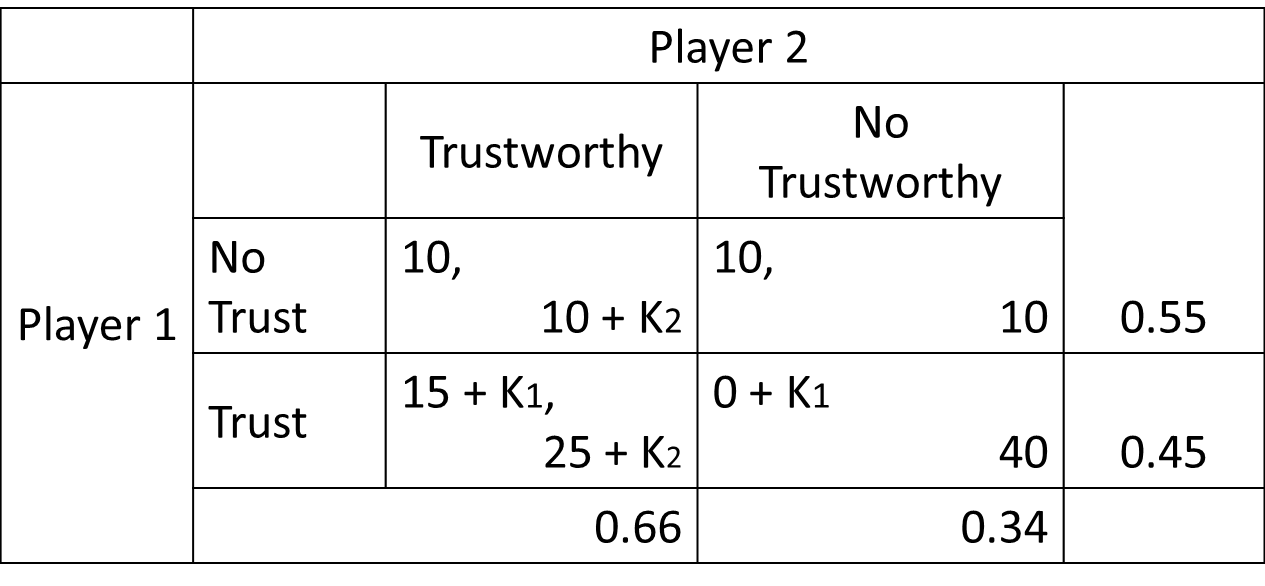}
  \caption{The trust game in normal form with Kantian value included.  }\label{fig:TrustKant}
\end{figure}

Figure \ref{fig:TrustKant} presents the trust game in normal form game, with the Kantian values of Player 1 (denoted as $K_1$) and Person 2 (denoted as $K_2$) added. In addition, the outcome strategy probability of Person 1 and 2 from experiment data are shown in the last column and raw, respectively. Supposing the given strategy profile is the Nash equilibrium of this 2 by 2 game, we can calculate the Kantian values of the two roles respectively, obtain that
\begin{center}
$K_1 \simeq \$0$ and  $K_1 \simeq \$7$
\end{center}
The experiments have shown that human subjects have incentive to benefit the recipients, from which the Kantian value can be taken to restrict AI ethics.
And experiments, such as dictators game, public goods game, centipede game ,etc., have also shown that human subjects maintain a high degree of consistency across multiple versions of the similar game \cite{Fehr1999A,Camerer2003}.
The consistency suggests that it is possible to find the regulation of the Kantian values over various human subjects game \cite{Fehr1999A}.

\section{Technology for testing AI ethics}

Algorithm agent is a typical AI product. To test whether such a agent aligned human value, we need a practical  technology methods. To this end, we use a laboratory experiment of iterated prisoner dilemma games, in which AI algorithm agents competes with human subjects. We will illustrate how to distinguish the ethical algorithm and the unethical ones with mathematical representation.

\subsubsection{Iterated prisoner's dilemma}
\begin{figure}[tbh]
  \centering
     \includegraphics[scale = 0.08]{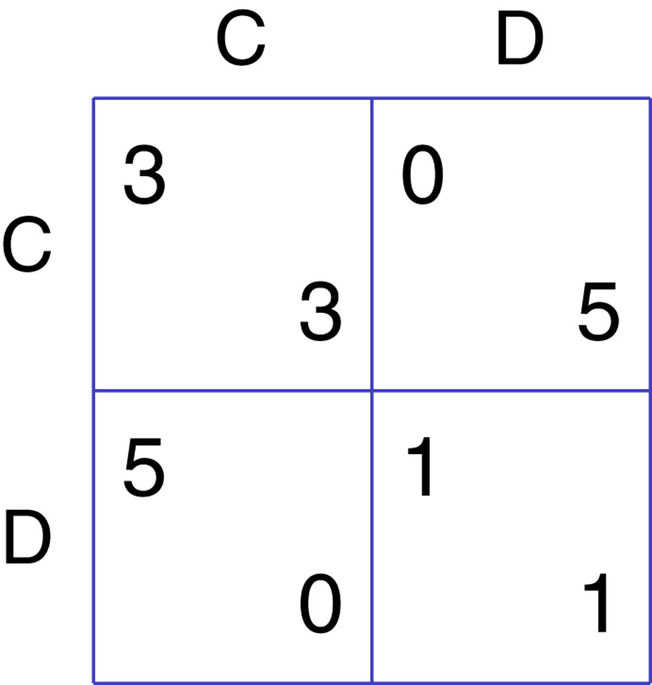}
  \caption{Payoff matrix of prisoner dilemma.
   }\label{fig:IPDPayoffMatrix}
\end{figure}

Promoting social cooperation is an important challenge. The iterated prisoner's dilemma (IPD) has been widely studied as the canonical game theoretic framework representing this issue \cite{Axelrod1984The}. In a one-shot two-person prisoner's dilemma, there are two pure strategies: cooperate and defect. Each player receives R if they mutually cooperate; each player receives $P$ if they mutually defect; if one player cooperates and the other defects, the defector receives $T$ and the cooperator receives S, where $T>R>P>S$ guarantees that in this game the commonly used solution concept Nash equilibrium is mutual defection, while $2R>T+S$ implies that mutual cooperation is actually the socially best outcome. A typical specification suggested by Axelrod  (1984) is $R=3, T = 5, S=0, P = 1$, shown in Figure \ref{fig:IPDPayoffMatrix}.

\subsubsection{Traditional human ethics in IPD}
Robert Axelrod in his book The Evolution of Cooperation \cite{Axelrod1984The} 
 reports a tournament he organized of the IPD game, whose participants are from academic colleagues all over the world.
 Participants were asked to devise algorithms to play IPD game against all the other participants one by one, 500 rounds for each, with the memory of all the previous round against the current opponent.
 The score is the sum of the payoffs of all the rounds.
It is discovered that for fixed-match repeated game,
the algorithm with greedy (unethical) strategies tends to fail in the long run while the one with more altruistic (ethical) strategies won.
He used this to show a possible mechanism for the evolution of altruistic (ethical) behaviors from mechanisms that are initially purely selfish, by natural selection.
By analysing the top-scoring strategies, Axelrod stated several surveyable (and then ethical) characteristics.

\begin{enumerate}
  \item  Nice:
  it will not defect before its opponent does.
  \item Retaliating:
    not be a blind optimist. It must sometimes retaliate.
    Cooperation without retaliating could lead to being exploited ruthlessly.
  \item Forgiving:
    must also be forgiving. Though players will retaliate, they
    need to recover cooperation sometimes
    from long runs of revenge and counter-revenge, to maximize points.
  \item  Non-envious:
    The last quality is being non-envious, that is not striving to score more than the opponent.
\end{enumerate}
We use these as ethics reference to test an algorithm by linguistic analysis (results see Fig \ref{fig:ZDoutcome2}).

\subsubsection{Tested samples}  
\begin{figure}[tbh]
  \centering
      \includegraphics[scale = 0.25]{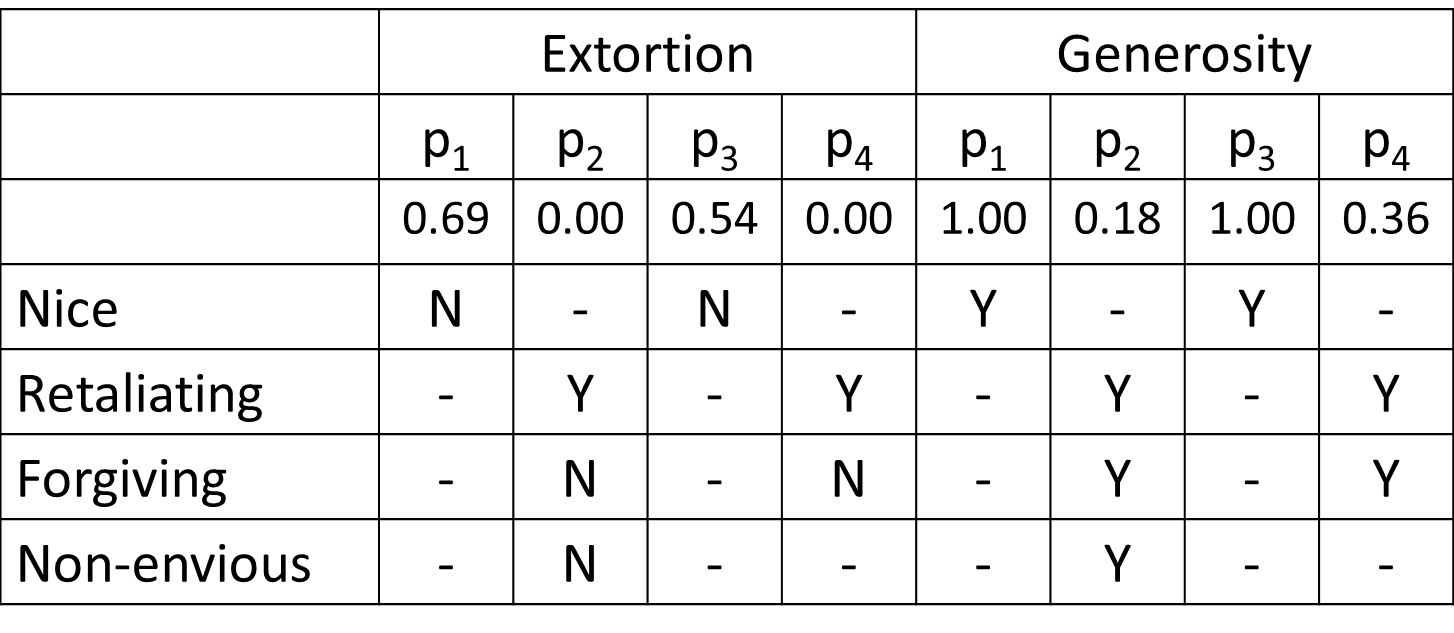}
   \caption{Two tested AI algorithm, named as Extorter and Generosity.
   The values in 3nd raw is assigned for the two Algorithm.
   'Y', 'N' and '-' refer to the ethical requirement satisfied, not satisfied and indistinguishable, respectively, which evaluated by linguistic analysis referring to traditional human ethics suggested by Axelrod mentioned above.}\label{fig:ZDoutcome2}
\end{figure}

We use zero-determinant (ZD) strategies algorithm as the tested samples,
 which allows a player to unilaterally enforce a linear relationship between his payoff and that of his opponent\cite{Press2012Iterated}. A ZD strategy is described by the probabilities of cooperation given the four possible outcomes of the previous round: $p$=$(p_1, p_2, p_3, p_4)$, where $p_i$, $i\! \in\!(1, 2, 3, 4)$ is the probability of cooperation given the previous outcomes $CC, CD, DC$ and $DD$, respectively. Two ZD algorithms, named as Extorter and Generosity \cite{Wang2016Extortion}, are specified (shown in \ref{fig:ZDoutcome2}) to demonstrate the ethical testing.\\

\subsubsection{Testing technology} 
\begin{figure}[tbh]
  \centering
\includegraphics[scale = 0.19]{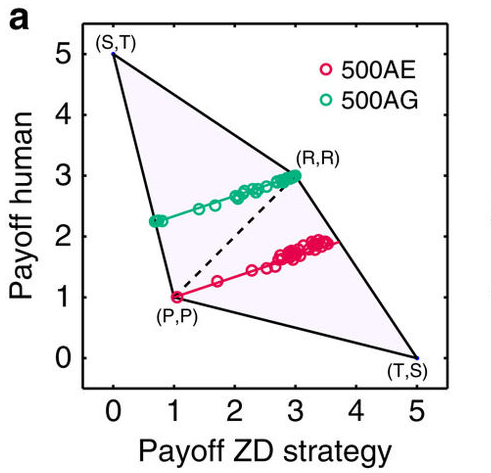}
  \caption{Experimental scores and theoretical prediction. The red (green) line corresponds to the theoretical outcome space of Extortion (Generosity) algorithm. Each open circle indicates a pair of scores of AI and human in the experiments.}\label{fig:ZDoutcome}
\end{figure}

Original outcome space is a quadrilateral zone grayed in Fig. \ref{fig:ZDoutcome}. However, when the AI agent uses a ZD algorithm, outcome space is limited and collapsed to the red or green line.
We can distinguish the ethics of the two algorithm with the lines.
As results, Generosity (green line in Fig \ref{fig:ZDoutcome}) is ethical and the Extortion (red line in Fig. \ref{fig:ZDoutcome}) is unethical. Explanation is as follows.
For Extorter, its score $s_e$ and its human co-player's score $s_{he}$ satisfy $$\frac{s_{he}  -1}{s_e - 1} = \frac{1}{3},$$ illustrated as red line in Fig \ref{fig:ZDoutcome}. While the minimum scores for both Extorter and the human co-player are  $s_e^{min}$ = $s_{he}^{min}$ = 1, while the maximum scores
are $s_e^{max}$ = 3.727 and $s_{he}^{max}$ = 1.907 which is unfair definitively and unethical.

For Generosity, its score  $s_g$ and its human co-player's $s_{hg}$ score satisfy $$\frac{3- s_{hg}}{3 - s_g} = \frac{1}{3},$$ illustrated as green line in Fig \ref{fig:ZDoutcome}. The maximum scores for both are 3,
 which is fair and efficient, and then ethical.

%

\section{Related works}
 In the tree chart of AI ethics shown in Fig. \ref{fig:TreeChart1}, on the decision making branch, the cooperative game interaction has been well studied, (e.g., Russell and his colleagues \cite{hadfieldmenell2016cooperative}).
 This work is on the branch of non-cooperation game.

\begin{figure}[tbh]
  \centering
      \includegraphics[scale = 0.26]{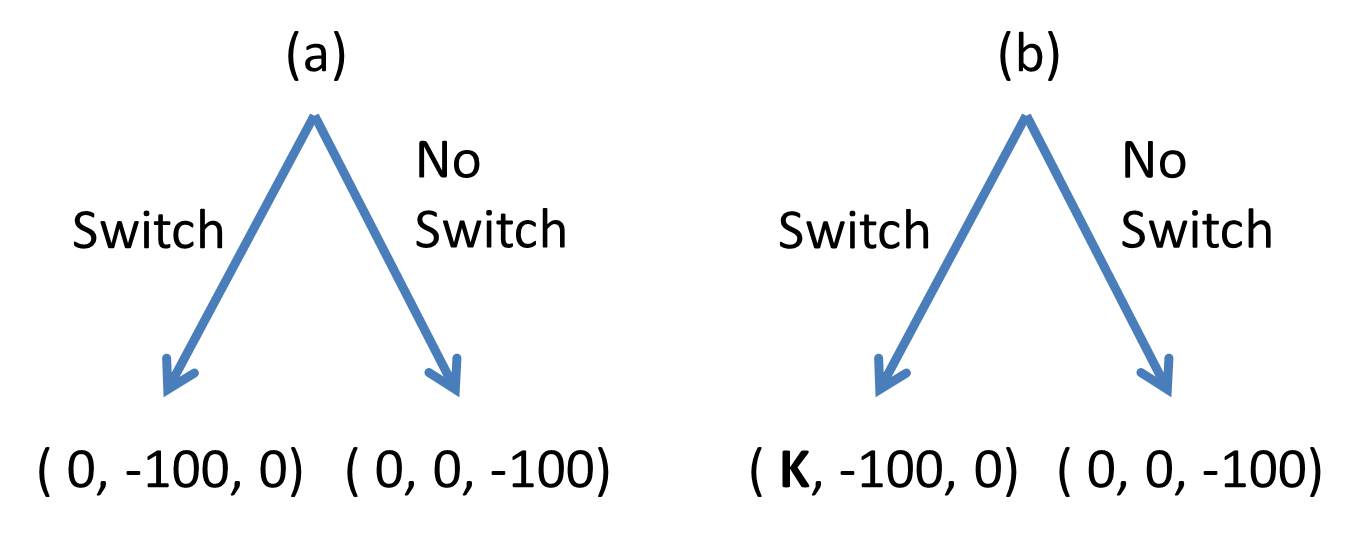}
  \caption{Comparison of two paradigms. (a) Previous presentation, how to include the player 2 and player 3's reward was not defined, and solution is unknown. (b) In our presentation , the Kantian value of an action is specified and the solution is computable. }\label{fig:Ytrain}
\end{figure}

We have introduced a mathematical representation, which differs from previous paradigms which can only include Utilitarian  
by including Kantian too. So, the game theory technology for
the solution concept
can be applied only dependent on the outcome space.
In previous paradigm \cite{Conitzer2017Moral}, shown in Fig. \ref{fig:Ytrain}(a), the reward to switch or not is same (both are 0) which put the agent in dilemma. On the contrary, in our presentation shown in Fig. \ref{fig:Ytrain}(b), the reward to switch or not is not same, in which if switch action is Kantian wrong, $K < 0$, and the solution is definitely not to switch.

For the technology of taking human value from experiment practically, we have illustrated an approach to take the human value (trust and trustworthy) from human subjects experiment. These parameters can be transit to AI in similar non-cooperation game condition, avoiding AI to be too tough.
Though it is well known that there exists the human value(e.g., fairness, justices) in the experimental game theory \cite{Crawford2002Introduction,Fehr1999A,Camerer2003}, our approach is the first to take the human value quantitatively
as AI ethics controlling parameter for AI design.

On AI ethics testing technology, in an IPD game, we have demonstrated how to test an AI algorithm being ethical or unethical by analyzing the outcome space (shown in Fig. \ref{fig:ZDoutcome}). Our method is a quantitative method which is more practical and accurate than the linguistic analysis method. Appearing as check list (as shown in Figure \ref{fig:ZDoutcome2}), the linguistic analysis method could lead to arguable results, however, it is wildly used in IEEE P7000-series ongoing AI ethics industry technology standard developing nowadays \cite{Raja2017The}.


\section{Discussion and conclusion}
Our work in this paper provides a quantitative approach, whose data is derived from human subject experiments of game theory. As mentioned above, in order for AI to align human value, a rigorous scientific method is desired. That means, both quantitative experiment and computational methods are both necessary.

In the experimental side, experimental behavior science is well studied \cite{Plott2008}. Many models, like public goods game, hawk and dove game, chicken game and so on, have been developed to well describe various competing conditions. Thus taking human values from these experiments is possible. Experiments of neuroscience can also be references, from which human values on justices, altruism, fairness and so on can be taken (e.g., \cite{Fehr2007Social}). Analysing big data from real life human behaviors, and data from survey \cite{bonnefon2016the} or voting on moral dilemma method are valuable.

In the computational side, the merged neural systems, higher degree morphological agent \cite{Mathews2017Mergeable}, collective (emergent) effect of multi AI agents are all needed for further investigations. Computational methods are not only meaningful for individual ethical AI, but also for AI social impacts (e.g., workforce displacement, economics inequality, etc). In this direction, social computing methods \cite{Wang2007Social}, e.g., agent based simulation, multi-AI-Human interaction simulation, normative multi agent simulations) are desired too. Nevertheless, we also hope continuous development of linguistic logical computing methods \cite{sowa1999knowledge} can help the commendation between the arguments among the stakeholder of various laws, regulations and cultures, which can help the establishment of the industry technology standard (e.g., IEEE P7000 series).

 \bibliographystyle{aaai}
 \bibliography{bibsample}

\begin{thebibliography}{}

\bibitem[\protect\citeauthoryear{Allen, Wallach, and Smit}{2006}]{Allen2006Why}
Allen, C.; Wallach, W.; and Smit, I.
\newblock 2006.
\newblock Why machine ethics?
\newblock {\em IEEE Intelligent Systems} 21(4):12--17.

\bibitem[\protect\citeauthoryear{Axelrod}{1984}]{Axelrod1984The}
Axelrod, R.
\newblock 1984.
\newblock {\em The Evolution of Cooperation Basic Books}.
\newblock Basic Books,.

\bibitem[\protect\citeauthoryear{Bonnefon, Shariff, and
  Rahwan}{2016}]{bonnefon2016the}
Bonnefon, J.; Shariff, A.~F.; and Rahwan, I.
\newblock 2016.
\newblock The social dilemma of autonomous vehicles.
\newblock {\em Science} 352(6293):1573--1576.

\bibitem[\protect\citeauthoryear{Camerer and Foundation}{2003}]{Camerer2003}
Camerer, C., and Foundation, R.~S.
\newblock 2003.
\newblock {\em Behavioral game theory: Experiments in strategic interaction},
  volume~9.
\newblock Princeton University Press Princeton, NJ.

\bibitem[\protect\citeauthoryear{Chatila \bgroup et al\mbox.\egroup
  }{2017}]{Raja2017The}
Chatila, R.; Firth-Butterflied, K.; Havens, J.~C.; and Karachalios, K.
\newblock 2017.
\newblock The ieee global initiative for ethical considerations in artificial
  intelligence and autonomous systems [standards].
\newblock {\em IEEE Robotics \& Automation Magazine} 24(1):110--110.

\bibitem[\protect\citeauthoryear{Conitzer \bgroup et al\mbox.\egroup
  }{2017}]{Conitzer2017Moral}
Conitzer, V.; Sinnottarmstrong, W.; Borg, J.~S.; Deng, Y.; and Kramer, M.
\newblock 2017.
\newblock Moral decision making frameworks for artificial intelligence.

\bibitem[\protect\citeauthoryear{Crawford}{2002}]{Crawford2002Introduction}
Crawford, V.~P.
\newblock 2002.
\newblock Introduction to experimental game theory.
\newblock {\em Journal of Economic Theory} 104(1):1--15.

\bibitem[\protect\citeauthoryear{Fehr and Camerer}{2007}]{Fehr2007Social}
Fehr, E., and Camerer, C.~F.
\newblock 2007.
\newblock Social neuroeconomics: the neural circuitry of social preferences.
\newblock {\em Trends in Cognitive Sciences} 11(10):419--427.

\bibitem[\protect\citeauthoryear{Fehr and Schmidt}{1999}]{Fehr1999A}
Fehr, E., and Schmidt, K.~M.
\newblock 1999.
\newblock {\em A Theory of Fairness, Competition, and Cooperation}.
\newblock University of Munich, Department of Economics.

\bibitem[\protect\citeauthoryear{Friedman \bgroup et al\mbox.\egroup
  }{2013}]{Friedman2013Value}
Friedman, B.; Jr, P. H.~K.; Borning, A.; and Huldtgren, A.
\newblock 2013.
\newblock {\em Value Sensitive Design and Information Systems}.

\bibitem[\protect\citeauthoryear{Hadfieldmenell \bgroup et al\mbox.\egroup
  }{2016}]{hadfieldmenell2016cooperative}
Hadfieldmenell, D.; Dragan, A.~D.; Abbeel, P.; and Russell, S.~J.
\newblock 2016.
\newblock Cooperative inverse reinforcement learning.
\newblock {\em neural information processing systems}  3909--3917.

\bibitem[\protect\citeauthoryear{Kuhn}{2004}]{Kuhn2004Reflections}
Kuhn, S.~T.
\newblock 2004.
\newblock Reflections on ethics and game theory.
\newblock {\em Synthese} 141(1):1--44.

\bibitem[\protect\citeauthoryear{Lasota, Fong, and Shah}{2017}]{MITsafety2017}
Lasota, P.~A.; Fong, T.; and Shah, J.~A.
\newblock 2017.
\newblock A survey of methods for safe human-robot interaction.
\newblock {\em Foundations and Trends in Robotics} 5(3):261--349.

\bibitem[\protect\citeauthoryear{Letchford, Conitzer, and
  Jain}{2008}]{Letchford2008An}
Letchford, J.; Conitzer, V.; and Jain, K.
\newblock 2008.
\newblock An "ethical" game-theoretic solution concept for two-player
  perfect-information games.
\newblock In {\em Internet and Network Economics, International Workshop, Wine
  2008, Shanghai, China, December 17-20, 2008. Proceedings},  696--707.

\bibitem[\protect\citeauthoryear{Mathews \bgroup et al\mbox.\egroup
  }{2017}]{Mathews2017Mergeable}
Mathews, N.; Christensen, A.~L.; O'Grady, R.; Mondada, F.; and Dorigo, M.
\newblock 2017.
\newblock Mergeable nervous systems for robots.
\newblock {\em Nature Communications} 8(1).

\bibitem[\protect\citeauthoryear{Osborne and Rubinstein}{1994}]{Osborne1994A}
Osborne, M.~J., and Rubinstein, A.
\newblock 1994.
\newblock {\em A course in game theory /}.
\newblock MIT Press,.

\bibitem[\protect\citeauthoryear{Plott and Smith}{2008}]{Plott2008}
Plott, C., and Smith, V.
\newblock 2008.
\newblock {\em Handbook of experimental economics results}.
\newblock North-Holland.

\bibitem[\protect\citeauthoryear{Press and Dyson}{2012}]{Press2012Iterated}
Press, W.~H., and Dyson, F.~J.
\newblock 2012.
\newblock Iterated prisoner's dilemma contains strategies that dominate any
  evolutionary opponent.
\newblock {\em Proceedings of the National Academy of Sciences of the United
  States of America} 109(26):10409.

\bibitem[\protect\citeauthoryear{Smith and Wilson}{2014}]{Smith2014Fair}
Smith, V.~L., and Wilson, B.
\newblock 2014.
\newblock Fair and impartial spectators in experimental econ.
\newblock {\em Review of Behavioral Economics} 1(1-2):1--26.

\bibitem[\protect\citeauthoryear{Sowa}{1999}]{sowa1999knowledge}
Sowa, J.~F.
\newblock 1999.
\newblock Knowledge representation: logical, philosophical and computational
  foundations.
\newblock {\em Computational Linguistics} 27(2):286--294.

\bibitem[\protect\citeauthoryear{Wang \bgroup et al\mbox.\egroup
  }{2007}]{Wang2007Social}
Wang, F.~Y.; Carley, K.~M.; Zeng, D.; and Mao, W.
\newblock 2007.
\newblock Social computing: From social informatics to social intelligence.
\newblock {\em IEEE Intelligent Systems} 22(2):79--83.

\bibitem[\protect\citeauthoryear{Wang \bgroup et al\mbox.\egroup
  }{2016}]{Wang2016Extortion}
Wang, Z.; Zhou, Y.; Lien, J.~W.; Jie, Z.; and Xu, B.
\newblock 2016.
\newblock Extortion can outperform generosity in the iterated prisoner's
  dilemma.
\newblock {\em Nature Communications} 7:11125.

\end{thebibliography}

\end{document}